\documentclass[10pt,twocolumn,letterpaper]{article}

\usepackage{iccv}
\usepackage{times}
\usepackage{epsfig}
\usepackage{graphicx}
\usepackage{amsmath}
\usepackage{amssymb}
\usepackage{kotex}
\usepackage{xcolor}

\usepackage[pagebackref=true,breaklinks=true,letterpaper=true,colorlinks,bookmarks=false]{hyperref}

\iccvfinalcopy 

\def\iccvPaperID{****} 

\ificcvfinal\fi

\begin{document}

\title{Instruction-tuned Self-Questioning Framework for Multimodal Reasoning}


\author{You-Won Jang\\
Seoul National University\\
ywjang@bi.snu.ac.kr\\
\and
Yu-Jung Heo\\
KT\\
yj.heo@kt.com\\
\and
Jaeseok Kim\\
KT\\
jaeseok.kim@kt.com\\
\and
Minsu Lee\\
Seoul National University\\
minsue.lee@gmail.com\\
\and
Du-Seong Chang\textsuperscript{\rm {$\ast$}}\\
KT\\
duseong.chang@gmail.com\\
\and
Byoung-Tak Zhang\textsuperscript{\rm {$\ast$}}\\
Seoul National University\\
btzhang@bi.snu.ac.kr\\
}


\maketitle
\ificcvfinal\thispagestyle{empty}\fi

\newcommand{\MODELNAME}{\textcolor{black}{SQ-InstructBLIP}}

\begin{abstract}
   The field of vision-language understanding has been actively researched in recent years, thanks to the development of Large Language Models~(LLMs). However, it still needs help with problems requiring multi-step reasoning, even for very simple questions. Recent studies adopt LLMs to tackle this problem by iteratively generating sub-questions and answers. However, there are disadvantages such as 1) the fine-grained visual contents of images are not available using LLMs that cannot read visual information, 2) internal mechanisms are inaccessible and difficult to reproduce by using black-box LLMs. To solve these problems, we propose the SQ(Self-Questioning)-InstructBLIP, which improves inference performance by generating image-aware informative sub-questions and sub-answers iteratively. The \MODELNAME, which consists of a Questioner, Answerer, and Reasoner that share the same architecture. Questioner and Answerer generate sub-questions and sub-answers to help infer the main-question, and Reasoner performs reasoning on the main-question considering the generated sub-question information. Our experiments show that the proposed method \MODELNAME, which uses the generated sub-questions as additional information when solving the VQA task, performs more accurate reasoning than the previous works.
\end{abstract}

\section{Introduction}



\begin{figure}[t]
\begin{center}
   \includegraphics[width=1.\linewidth]{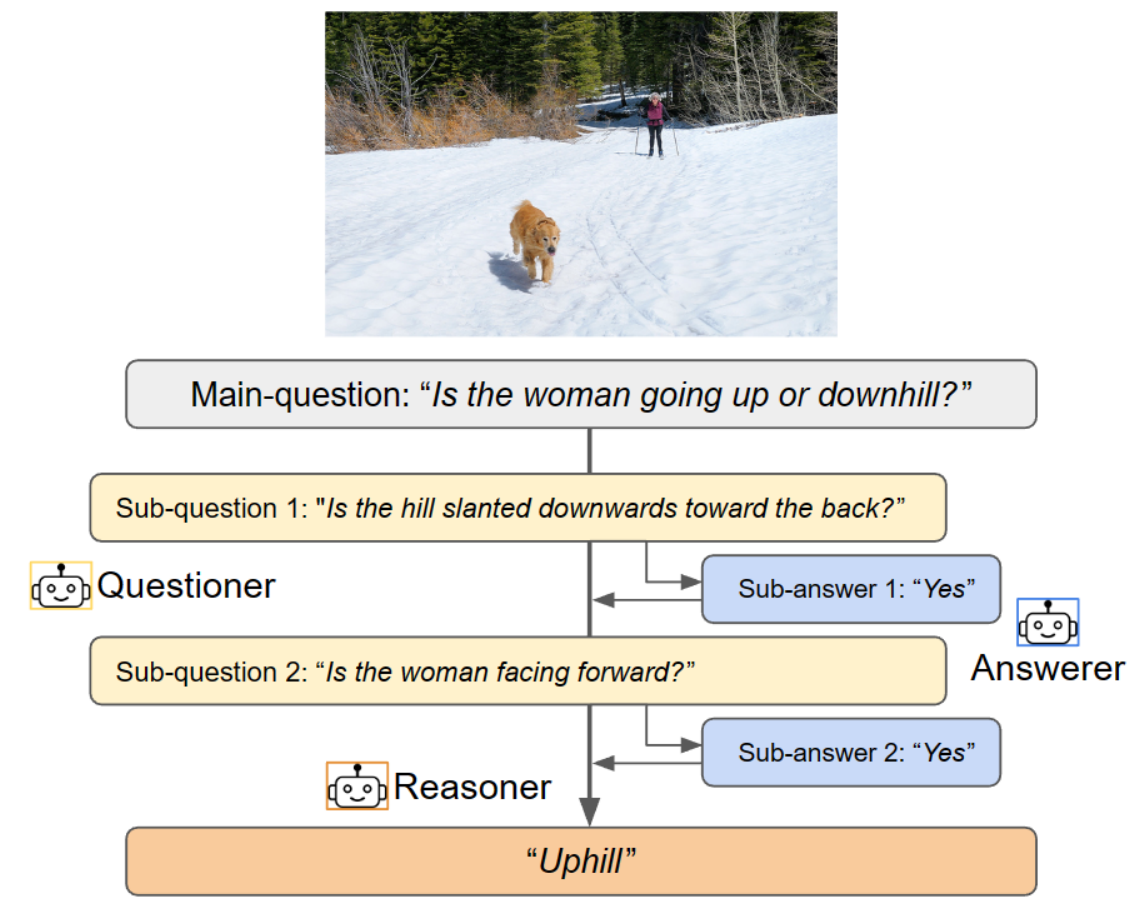}
\end{center}
   \caption{Example of a self-questioning process for multi-step multimodal reasoning.}
\label{fig:figure01}
\end{figure}

In the realm of vision-language understanding, pre-trained models have yielded remarkable accomplishments across various downstream tasks, attributed mainly to the potency of super-sized language models. However, these models have encountered difficulties when confronted with tasks necessitating sequential reasoning steps, in contrast to single-step processes. Despite addressing seemingly uncomplicated inquiries, they tend to conduct inferences in a single step, while the underlying problem often mandates a series of steps. For example, as illustrated in Figure \ref{fig:figure01}, discerning whether the woman is ascending or descending involves several internal deliberations: evaluating the hill's relative height, the woman's gaze direction, and subsequently arriving at a conclusion—``uphill". This instance underscores the necessity for multi-step reasoning, even for seemingly straightforward questions.

Building upon this concept, recent studies have proposed the self-questioning scheme for multi-step multimodal reasoning. Uehara \etal\cite{9857023} adopt a strategy of generating sub-questions via a visual question generation (VQG) model. This model is trained with an Info-score module to evaluate the efficacy of the generated questions. However, a limitation arises from generating only one question, leading to insufficient additional information. On the other hand, You \etal\cite{you2023idealgpt} and Qi \etal\cite{qi2023art} suggest generating multiple sub-questions utilizing ChatGPT~\cite{chatgpt}. However, since it relies solely on language-based models (\ie LLMs), these have a clear limitation in obtaining fine-grained information about the given image. Furthermore, these modules depend on ChatGPT, which is difficult to reproduce.

In this paper, to resolve these problems, we propose \MODELNAME, which iteratively generates sub-questions and sub-answers to answer a given main-question utilizing a vision-language model (VLM) rather than a language-only model. 
Our contributions are as follows:
\begin{itemize}
    \item We introduce a novel method to iteratively generate sub-questions that ask for diverse contents, so that as much information as possible can be obtained from the sub-questions.
    \item We propose a method using VLM that can utilize the fine-grained contents of the image for all modules of our architecture, allowing for more informative and accurate sub-questions and sub-answers.
    \item We prove that using the sub-questions as additional information improves the performance of the VQA task, and show that the more accurate and the more sub-questions we generated, the higher the performance.
\end{itemize}

\begin{figure}[t]
\begin{center}
    \includegraphics[width=1.\linewidth]{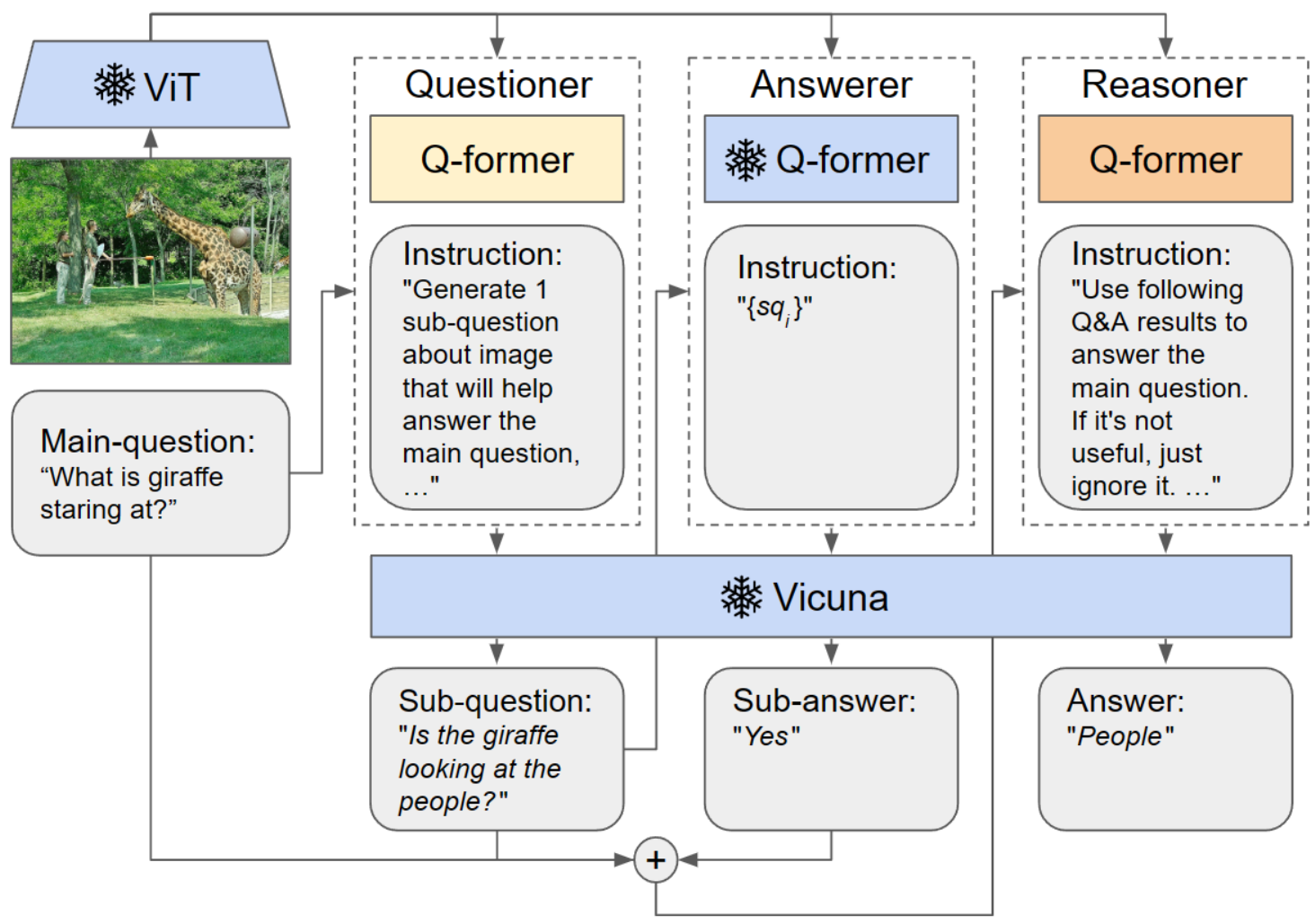}
\end{center}
   \caption{An overview of the proposed \MODELNAME~Framework.}
\label{fig:Framework}
\end{figure}



\section{Method}

\begin{figure}[t]
\begin{center}
    \includegraphics[width=0.8\linewidth]{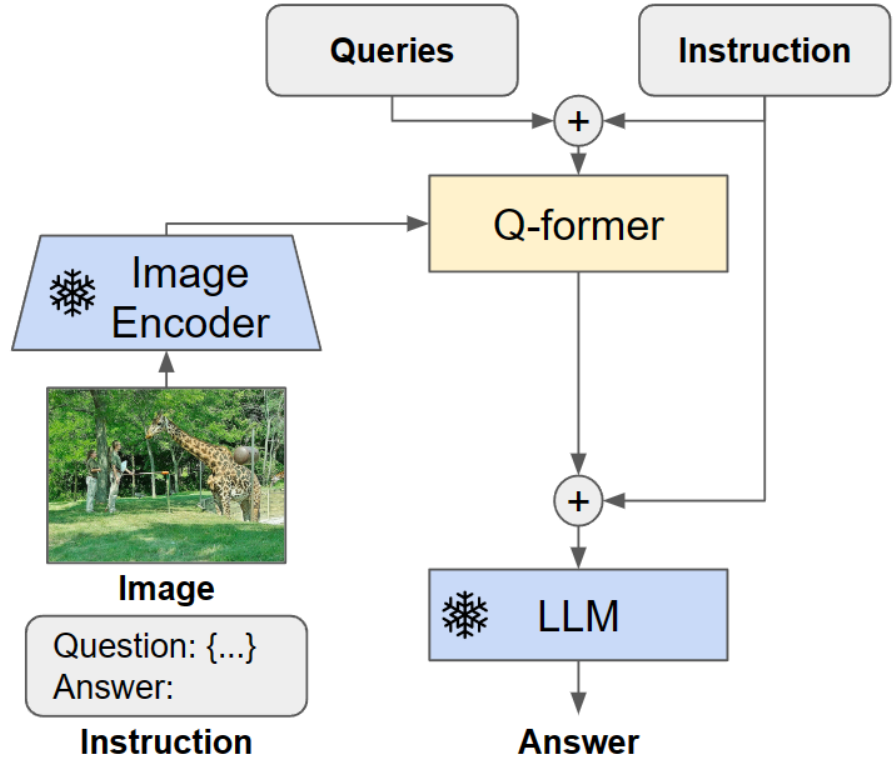}
\end{center}
   \caption{Simplified InstructBLIP structure consisting of image encoder, Q-former, and LLM.}
\label{fig:InstructBLIP}    
\end{figure}

\MODELNAME~consists of three components such as 1) a \textbf{Questioner} that generates sub-questions, 2) an \textbf{Answerer} that answers the sub-questions, and 3) a \textbf{Reasoner} that performs reasoning about the main-question based on the generated sub-questions and sub-answers (see Figure \ref{fig:Framework}). We emphasize that the proposed framework is module-agnostic; thereby, \textbf{Questioner}, \textbf{Answerer} and \textbf{Reasoner} can be instantiated based on any VLM model. 
In this work, we employ InstructBLIP \cite{dai2023instructblip}, which is a representative instruction-tuned VLM model as a base model. 
As shown in Figure \ref{fig:InstructBLIP}, InstructBLIP trains only the Q-Former module while freezing the image encoder and LLM. Following the previous work, we also train the Q-former for \textbf{Questioner} and \textbf{Reasoner}. 

In the subsequent section, $q$ represents the main-question, while $sq_i$ and $sa_i$ denote the $i$-th generated sub-question and its corresponding sub-answer, respectively.


\subsection{Questioner}
\textbf{Questioner} is a sequence-to-sequence VLM-based model that generates informative sub-questions given a main-question, trained by instruction-tuning. In order to obtain more useful information to answer the main-question, we carefully design the \textbf{Questioner}'s instruction as follows. 


First, we design the instruction to generate the first sub-question, $sq_1$, given $q$ with the prompt: 
\textit{Generate 1 sub-question about image that will help answer the main-question, when main-question is `$\{q\}$'}. When creating the second and later, $sq_i, \ i\ge2$, we add an instruction to ask about different information than the previously created sub-questions and used it as input as follows: \textit{Create a question that asks about different information than the following questions. $\{sq_1\}, ..., \{sq_{i-1}\}.$}

\subsection{Answerer}

\textbf{Answerer} is a model that infers an answer to a given question, the same as a general visual question answering model. While doing self-questioning, getting the correct answers to sub-questions is essential.
It is best to use an oracle (\eg human) that can accurately answer sub-questions. However, asking humans to answer all of the generated sub-questions is practically impossible, so we adopt a VLM as an \textbf{Answerer}. We utilize the most straightforward prompt using only a sub-question as follows: \textit{\{$sq_i$\}}


\subsection{Reasoner}
\textbf{Reasoner} is a model that infers the final answer based on all contexts when given a main-question, sub-questions, and sub-answers. We proceed with fine-tuning based on a VLM model since the existing VLM model has not been trained to perform complex inferences based on several relational facts.
The instruction of \textbf{Reasoner} is to refer to $(sq_i, sa_i)$ when answering the main-question: \textit{Use the following Q\&A results to answer the main-question. If it's not useful, just ignore it}. Then, the sub-question and sub-answer pair, and the main-question are appended as follows: \textit{\{$sq_1$\} \{$sa_1$\}. ...  \{$sq_i$\} \{$sa_i$\}. main-question}: \{$q$\} \textit{A}:

\section{Experiments}

\subsection{Datasets}

\textbf{VQA-Introspect} dataset \cite{selvaraju2020squinting} is a subsequent dataset of VQAv2~\cite{goyal2017making} dataset. It contains additional annotations of sub-questions and sub-answers to validate intermediate-level reasoning for answering main-question in the VQAv2 dataset. A total of 238k sub-questions and sub-answers on about 77k images are included in the dataset. Train split contains 38k images and 167k sub-questions for 56k reasoning questions. The validation split contains 17k images and 72k sub-questions for 22k Reasoning questions.



\textbf{A-OKVQA}~\cite{schwenk2022aokvqa} is an augmented successor of OK-VQA, a representative knowledge-based VQA benchmark. The questions in A-OKVQA require a diverse base of world knowledge, such as commonsense, visual concepts, and knowledge from textbooks. In this work, we utilize its validation split of 1.1k questions to validate \MODELNAME~in a zero-shot evaluation manner.

\subsection{Implementation details}

All three modules, \textbf{Questioner}, \textbf{Answerer}, \textbf{Reasoner}, utilize the same VLM structure, InstructBLIP-vicuna7b \cite{dai2023instructblip, vicuna}, as the base architecture. In the self-questioning framework, the image encoder (\ie, vision transformer \cite{dosovitskiy2021image}) and the language model are frozen as shown in Figure \ref{fig:Framework}. We initialize \textbf{Questioner} with the InstructBLIP-vicuna7b model, which is open to the public, and fine-tune the model under the sequence-to-sequence generation objective. We follow almost all of the hyper-parameter settings in the pre-trained models, except the batch size, warm-up step and gradient accumulation.
For \textbf{Answerer}, we use the InstructBLIP-vicuna7b without any additional training. Furthermore, we fine-tune the \textbf{Reasoner} in the same manner as for the \textbf{Questioner}. The ground-truth sub-questions and sub-answers are used to train the \textbf{Reasoner}, not the generated sub-question and answer pairs. During the fine-tuning phase for both the questioner and the reasoner, the training of 5 epochs necessitates approximately 8 hours, utilizing a single A100 (40GB) GPU.


\subsection{Baseline models}
We adopt the three baseline models as follows:
1) Uehara \etal\cite{9857023}
2) SOCRATIC \cite{qi2023art} and
3) InstructBLIP \cite{dai2023instructblip}. 
For a fair comparison, we utilize the baseline's prompt to be the same as the last part of the \textbf{Reasoner}'s prompt.


\section{Results}

\begin{figure*}
\begin{center}
    \includegraphics[width=1.0\textwidth]{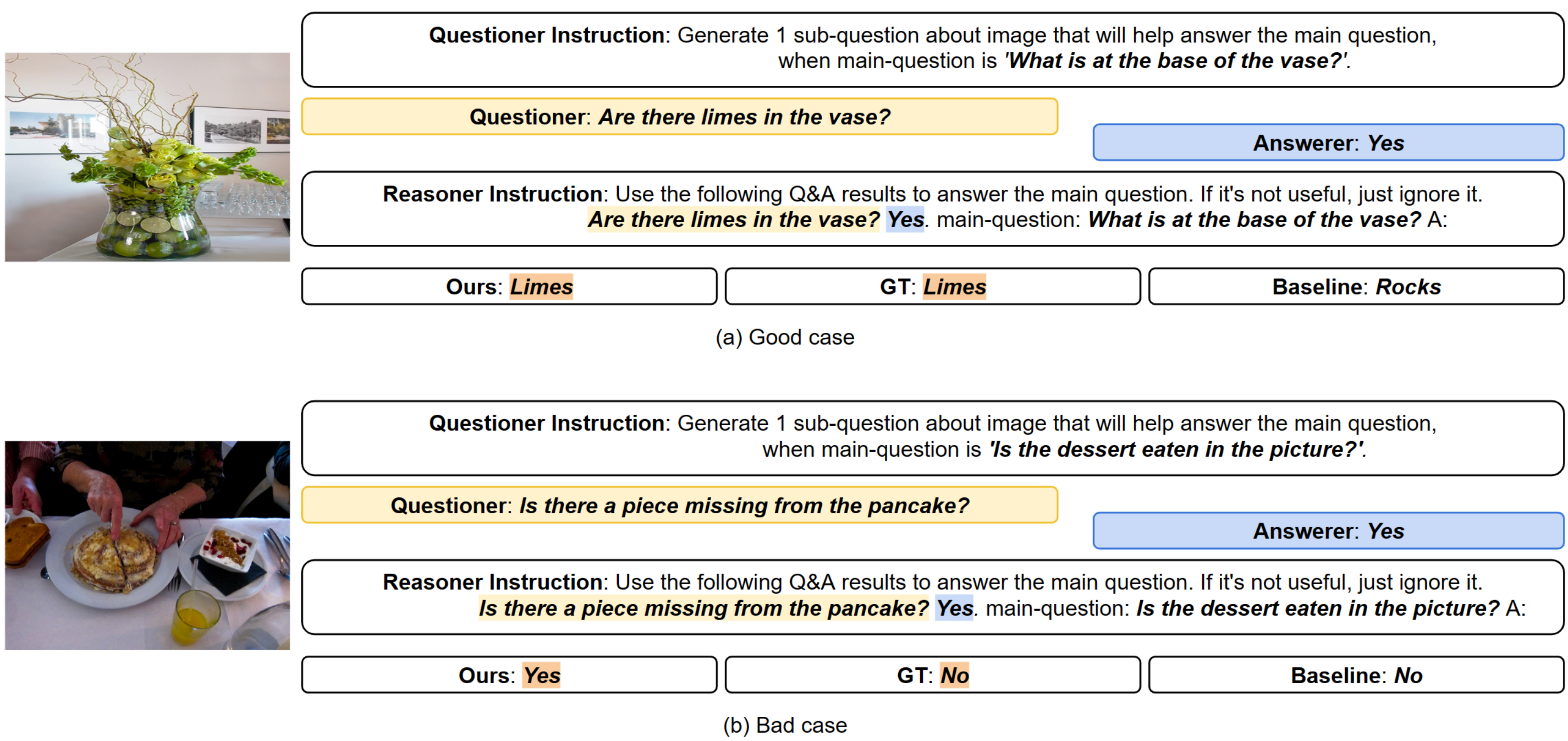}
\end{center}
   \caption{Qualitative Result: Generated informative sub-question and sub-answer. The sub-question and sub-answer have a background color of yellow and blue, respectively. (a) An example where the sub-question and sub-answer are generated appropriately and the overall reasoning is conducted accurately. (b) An example where the sub-question is appropriate, but the \textbf{Answerer} answers incorrectly and makes incorrect inferences. }
\label{fig:Qualitative}
\end{figure*}

\subsection{Quantitative results}
We conduct a self-questioning scheme for the visual question answering task on VQAv2 and A-OKVQA. As shown in Table \ref{tab:Result}, we first evaluate how the performance of the \textbf{Reasoner} changes when the sub-questions and sub-answers generated by the \textbf{Questioner} and \textbf{Answerer} are used as additional input context. As shown in the first section of the table, the self-questioning scheme improves the visual reasoning performance in both datasets. Next, we conduct additional experiments with ground-truth sub-QAs instead of generated sub-QAs. In this setting, \textbf{Reasoner} takes the ground-truth sub-QAs as input context and performs visual reasoning. As we can see in the second section (with ground-truth), the performance improves by about 11\% or more when the ground-truth sub-QAs are given as context in the same setting. Hereby, we highlight the effectiveness of the self-questioning scheme.

\begin{table}[t]
\begin{center}
\renewcommand\tabcolsep{3pt}
\begin{tabular}{lcc}
    \hline
    & VQA-Introspect & \multicolumn{1}{c}{A-OKVQA}  \\
    & Accuracy & MC\\ 
    \hline
    \textbf{with generated SubQAs} & & \\
    Uehara \etal \cite{9857023} & 77.12 & - \\
    InstructBLIP \cite{dai2023instructblip} & 85.53 & 72.75 \\ 
    \MODELNAME~(ours) & \textbf{86.84} & \textbf{73.28}\\ 
    \hline
    \textbf{with ground-truth SubQAs} & & \\
    Uehara \etal \cite{9857023} & 81.92 & - \\
    \MODELNAME~(ours) & \textbf{91.23} & - \\ 
    \hline
\end{tabular}
\end{center}
\caption{Accuracy of validation split of VQA-Introspect and A-OKVQA dataset. MC is an abbreviation for multi-choice accuracy. We generate 3 sub-QAs in this experiment.}
\label{tab:Result}
\end{table}

However, when open-ended evaluation rather than multiple-choice evaluation is performed on A-OKVQA, the performance drops about 4\% (56.04 $\rightarrow$ 60.96) compared to the baseline model (i.e., InstructBLIP).
There are two reasons for the performance degradation. First, the \textbf{Answerer} can infer other words that follow similar semantics as answers, which means \textbf{Answerer} generates open-ended answers (e.g., cell phone vs. mobile phone). 
Suppose an \textbf{Answerer} makes a correct inference but produces a synonym for a ground-truth answer, and the \textbf{Reasoner} that references it also creates a synonym as an answer. In that case, the direct-answer accuracy is evaluated as incorrect.  
Second, the sub-answers generated by the \textbf{Answerer} are often incorrect.
As shown in Figure \ref{fig:Qualitative}~(b), when \textbf{Questioner} asked \textit{``If the pancake was missing a piece?"}, the \textbf{Answerer} incorrectly answered \textit{``Yes"}. Due to the imperfect performance of the \textbf{Answerer}, the overall performance may be lower. This is evidenced by the much higher performance when using ground-truth subQAs. This issue might be alleviated by utilizing the more accurate \textbf{Answerer}.


\subsection{Qualitative results}
Figure \ref{fig:Qualitative} shows the main-question and the subsequent sub-questions and sub-answers generated by the \MODELNAME. As shown in the (a), when the main-question was \textit{``What is at the base of the vase?"}, \textbf{Questioner} generated \textit{``Are there limes in the vase?"} as a sub-question that could be an intermediate step in reasoning, and \textbf{Answerer} also correctly answered \textit{``Yes"}. Based on this result, \textbf{Reasoner} successfully inferred \textit{``Limes"}, unlike the baseline model.



\section{Ablation study}
In the ablation study, we conduct a series of experiments whereby a random selection of 1 out of every 10 samples (approximately 2.2k instances) was drawn from the validation split of the VQA-Introspect dataset. 
We report the change in the performance of visual reasoning according to the number of iterations of self-questioning.
In this setting, ground-truth sub-questions and answers are utilized to eliminate the influence of noise caused by the inaccurate \textbf{Questioner} or \textbf{Answerer}. 
As shown in Table \ref{tab:ablation_SQ_number}, the more sub-questions we create, the better the accuracy. However, since the time increases as the turn of self-questioning increases, we set the optimal number of sub-questions to $3$.


\begin{table}[t]
\begin{center}
\renewcommand\tabcolsep{4.5pt}
\begin{tabular}{ccccccc} 
    \hline
    \# of SQs & 0 & 1 & 2 & 3 & 4 & Max \\
    \hline
    Acc & 85.08 & 89.47 & 90.57 & 91.23 & 91.44 & 91.67 \\
    \hline
\end{tabular}
\end{center}
\caption{Accuracy of validation split of VQA-Introspect by number of sequentially-generated sub-questions~(SQs).}
\label{tab:ablation_SQ_number}
\end{table}


\section{Conclusion}

In this paper, we propose an Instruct-tuned Self-Questioning Framework~(\MODELNAME) to improve multimodal reasoning ability. We design the fully image-aware Questioner-Answerer-Reasoner system, which generates the informative sub-questions and answers to help answer the main-question and use these sub-QAs when performing the visual question reasoning. \MODELNAME~obtains helpful information to answer the main-question by generating effective sub-QAs iteratively. In our experiment, the proposed framework improves the performance of a strong baseline model, InstructBLIP, on open-ended question answering on the VQA-Introspect dataset and multi-choice question answering on the A-OKVQA dataset. 
In future work, we plan to apply the proposed framework to other multimodal reasoning tasks, such as visual commonsense reasoning or visual entailment. 

\section*{Acknowledgement}

This work was partly supported by KT Corp., the Institute of Information \& Communications Technology Planning Evaluation grants (2022-0-00951-LBA: 20\%, 2022-0-00953-PICA: 10\%, 2021-0-02068-AIHub: 5\%, 2021-0-01343-GSAI: 5\%), and the National Research Foundation of Korea grants (2021R1A2C1010970: 5\%, RS-2023-00274280: 5\%) funded by the Korean government.



{\small
\bibliographystyle{ieee_fullname}
\bibliography{egbib}
}

\end{document}